\documentclass{ifacconf}

\usepackage{graphicx}      
\usepackage{natbib}        
\usepackage{amsmath}
\begin{document}
\begin{frontmatter}

\title{Hierarchically Decentralized Heterogeneous Multi-Robot Task Allocation System}


\author[First]{Sujeet Kashid} 
\author[First]{Ashwin D. Kumat} 

\address[First]{Department of Mechanical and Materials Engineering, University of Cincinnati, Cincinnati, OH, USA. {email: kashidsv@mail.uc.edu, kumatad@mail.uc.edu}}

\begin{abstract}                
With plans to send humans to the Moon and further, the supply of resources like oxygen, water, fuel, etc., can be satiated by performing In-Situ Resource Utilization (ISRU), where resources from the extra-terrestrial body are extracted to be utilized. These ISRU missions can be carried out by a Multi-Robot System (MRS). In this research, a high-level auction-based Multi-Robot Task Allocation (MRTA) system is developed for coordinating tasks amongst multiple robots with distinct capabilities. A hierarchical decentralized coordination architecture is implemented in this research to allocate the tasks amongst the robots for achieving intentional cooperation in the Multi-Robot System (MRS). $3$ different policies are formulated that govern how robots should act in the multiple auction situations of the auction-based task allocation system proposed in this research, and their performance is evaluated in a $2$D simulation called pyrobosim using ROS$2$. The decentralized coordination architecture and the auction-based MRTA make the MRS highly scalable, reliable, flexible, and robust.
\end{abstract}

\begin{keyword}
Multi-agent and Networked Systems, Aerospace, Robotics, Multi-Robot Task Allocation, Auctions, Multi-Robot System, Decentralized, NASA, Space Robotics Challenge Phase $2$, pyrobosim.
\end{keyword}

\end{frontmatter}

\section{Introduction}
As space exploration goes forward with plans to send humans to Earth's moon and beyond, planning for the provision of enough resources to support these longer-term manned missions on the lunar surface (such as water, oxygen, and fuel) becomes critical for discussion. One approach to satiate these needs is In-Situ Resource Utilization (ISRU), as suggested by \cite{[9]}. This can potentially be done through the use of fully autonomous robotic systems. Robots could be designed to operate autonomously for longer periods before, during, and after the presence of humans to handle these needs.


Many researchers working on Multi-Robot Systems (MRS) have reported some critical advantages of MRS over Single-Robot Systems. \cite{[10]} state that a Single-Robot System cannot be at more than one places at once, whereas an MRS can be. Thus, an MRS system has better spatial distribution which can be used where the robotic application is distributed in space, time, or functionality as concluded by \cite{[13]}. \cite{[11]} and \cite{[12]} mention that an MRS system can perform tasks more efficiently in terms of total time required to complete the task by operating multiple robots in parallel, and/or complete the task using less energy. \cite{[14]}, \cite{[15]} and \cite{[16]} unanimously agree that redundancy in robot capabilities or through data sharing in MRS potentially increases the robustness and fault-tolerance of the system. A single-robot system has multiple sensors and many complex mechanisms to perform multiple tasks, thus making the single-robot costly. Whereas, an MRS can have a lower cost by using a number of simple robots that perform simple tasks in larger quantities which are cheaper to build than a powerful single robot to perform multiple tasks, suggested by \cite{[13]} and \cite{[17]}. \cite{[17]} also mentions that an MRS can potentially have better system reliability, flexibility, scalability, and versatility than a single-robot system \cite{[17]}. In an MRS, robots with different capabilities can be combined together to perform complex tasks, and the system can be designed in such a way that the task will be completed even when there are robot failures.

Through cooperation, the true potential of multi-robot systems can be leveraged. Just like humans would efficiently share the workload based on their number and skills, robots working together by distributing the workload would effectively make use of robot teams and thus maximize their overall task performance. As mentioned by \cite{[7]}, robot teams that share information and utilize each others’ skills can truly be more than the sum of its parts. Researchers have formulated different models to achieve cooperation amongst a team of robots. \cite{[28]} and \cite{[29]} discuss the emergent cooperation model, which is mostly popular in the field of swarm robotics, where the robots in the team follow a simple set of rules, and a group-level cooperative behavior emerges due to the interaction of the robots with each other and the environment, without the robots explicitly working together. Although it is a simple and elegant model, \cite{[33]} concludes that it is difficult to design and ensure the required proper behavior will emerge. As discussed by \cite{[23]}, another model is the intentional model of cooperation, where robots cooperate explicitly and with purpose. This explicit cooperation is achieved mostly through task-related communication. The intentional cooperation model as compared to the emergent cooperation model is more suitable for real-world scenarios, as suggested by \cite{[7]}. Moreover, if robots can cooperate with each other intentionally, so can humans cooperate with robots intentionally. This, humans cooperating with multi-robot systems is the long-term goal of multi-robot research. The benefit of humans cooperating with the MRS performing ISRU mission is a plus for an intentional model of cooperation. In this research, intentional cooperation is used at the level of task allocation.

There are $3$ different coordination architectures for an MRS as discussed by \cite{[34]}. In a centralized coordination architecture, a central control unit manages all robots in the MRS and has global information, offering optimal task coordination. However, it is vulnerable to a single point of failure. Distributed coordination architecture, seen in swarm MRS, has no central control, with each robot acting individually based on local knowledge which provides the least optimal solutions. Yet, they are more robust than centralized architecture. Decentralized coordination architecture combines aspects of both, using local central control for small clusters of robots, balancing optimality and reliability.

\cite{[36]} identifies $4$ major steps in the workflow of multi-robot systems to achieve the goal: task decomposition, which refers to breaking down a complex goal into simpler tasks; coalition formation, meaning the formation of robot teams; task allocation, which involves assigning the earlier decomposed simpler tasks to robots or robot teams; and task execution, which entails performing the assigned tasks through a sequence of actions.

As defined by \cite{[37]}, Multi-Robot Task Allocation (MRTA) is the process of optimizing collective performance by deciding which robot should perform which task, and thus it achieves a common goal through coordinated behavior. Researchers have developed many different algorithms to allocate tasks to robots in an MRS; for example, auction-based approaches by \cite{[49]} and \cite{[50]}, distributed assignment algorithms by \cite{[51]} and stochastic methods by \cite{[52]}. The classical method of solving a linear assignment problem with a Hungarian method has also been used to solve the MRTA problem in \cite{[53]}. Other approaches, such as the Iterated assignment architectures for solving MRTA have been proposed as by \cite{[54]} and \cite{[55]}.

In this study, a hierarchically decentralized coordination architecture has been designed. An auction-based MRTA method, inspired from \cite{[7]} due to its simplistic beauty, is developed to allocate the decomposed tasks to the different robots for intentional cooperation in the MRS. Additionally, different policies are formulated for coalition formation and task allocation to improvise and compare the task allocation method's efficiency. As for the scout robots, a systematic search strategy for finding the resources on the moon is also suggested. The system is developed and evaluated using a $2$D simulator called the pyrobosim using ROS$2$.

\section{Problem Formulation}

The Space Robotics Challenge Phase $2$ (SRC2) announced by NASA in 2019 has been used as an example scenario to present the proposed high-level task allocation system for a heterogeneous multi-robot system (MRS). The NASA SRC2 provided $3$ different kinds of robots with different capabilities. The scout robot can explore the surface of the moon, and its depths using volatile sensors to detect the location of lunar resources that can be extracted. Thus, the scout robot is designed to perform the task of localizing the resource site. The excavator robot can navigate to the resource location, dig the resource volatiles from under the lunar regolith, and drop the resource volatile into the hauler robot’s bin, making excavating the resource its primary task. The hauler robots are designed to transport these collected resource volatiles from the resource excavation site to the processing plant. So the ISRU goal is decomposed into $3$ tasks, namely localizing resource sites, excavating resources, and transporting resources. Details of the challenge can be found on the \cite{[27]} website.

The NASA SRC$2$ simulation was based on a Gazebo $3$D simulation. Many low-level manipulations of the robots had to be done in that $3$D simulation. Since this research focuses on high-level robot cooperation behavior, using the $3$D simulation used in NASA SRC2 would have been overkill, and might have interfered with the variations that are being considered in this research for comparison. Therefore, a $2$D simulator called the pyrobosim, developed by \cite{[2]}, is used in this research, which only focuses on the high-level robot behavior in a simple environment. The idea is to test out different variations of the cooperation algorithm proposed in this research without worrying about low-level robot manipulations. The $2$D simulator pyrobosim, an open-source simulator based in ROS$2$ and Qt, is found to be suitable for this research.

\begin{figure}
\begin{center}
\includegraphics[width=8.4cm]{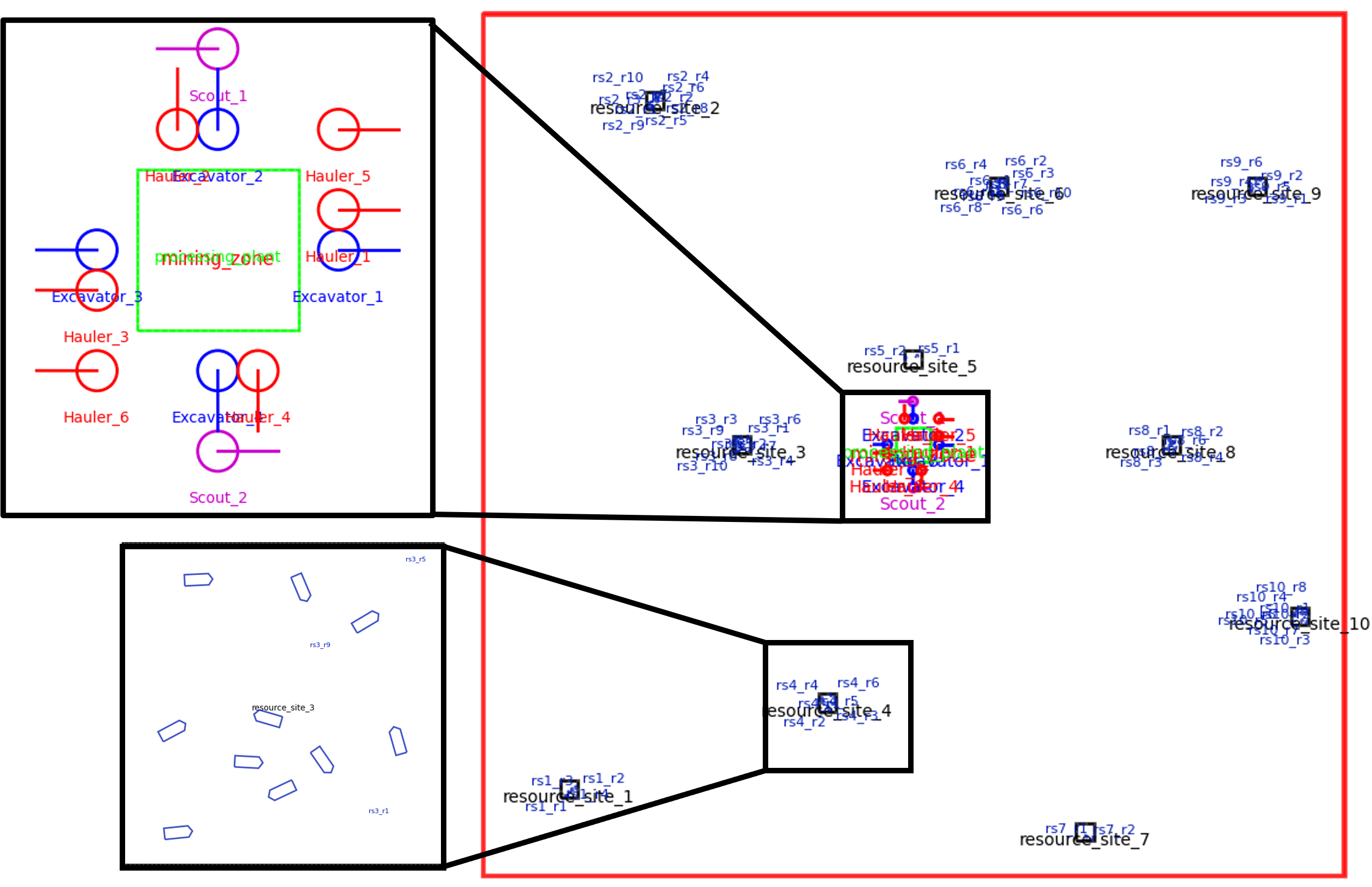}    
\caption{2D simulated area} 
\label{fig:2d_sim}
\end{center}
\end{figure}

In the 2D simulation of pyrobosim, 100 units x 100 units of area is considered for the lunar mining of the ISRU mission as shown in Fig. \ref{fig:2d_sim}. The Processing Plant is placed at the center of this lunar mining area, with the idea being all directions be equally accessible since the hauler robots must come to the Processing Plant to empty their bins from various locations within the lunar mining area. With the recognition of the task a particular type of robot has to perform, it can be seen that the scout robots have to move around the mining area a lot in search of resource sites. There will be many different resource sites found where the excavator robots have to keep digging for long times. To excavate at these resource sites, many excavator robots are needed, since an excavator will be working at a site for a long time to get all the resources out of it. Plenty of resource volatiles will be found at these sites that need to be transported from the site to the processing plant. With all the back-and-forth traveling between the site and the processing plant, many haulers will be needed to speed up the transportation. NASA SRC2 did not mention the size of the team of MRS or the quantities of each type or robot to be used. Hence, in this thesis, $2$ scout robots, $4$ excavator robots, and $6$ hauler robots are being used to model the MRS and develop and evaluate the MRTA system.

There are locations in this lunar mining area where volatile resources can be found. These locations will be henceforth referred to as resource sites. Once the scout robots find these resource sites, the excavator robots will come to the resource site to dig out the volatile resources. There will be varying amounts of volatiles found at a site. To quantify the amount of volatile resources, in this simulation, mineral objects are used as metaphors. Therefore, henceforth, one unit of volatile resources will be referred to as one mineral object. One mineral object is the maximum amount of volatile resource that a hauler robot can carry in its bin. These resource sites and mineral objects are depicted in the simulation as shown in Fig. \ref{fig:2d_sim}. There are 10 resource sites and 64 mineral objects randomly distributed amongst these 10 resource sites. The location of these resource sites in the lunar mining area, and the number of mineral objects present at each resource site are unknown to the robots. These are large enough numbers chosen for the purposes of the simulation and comparison in different scenarios. The scout robot has a scanning radius of 2.5 simulation meters around itself to find resource sites. The robots can determine their own progress and completion of tasks. Only high-level tasks are considered in this research. The robots can perform low-level tasks on their own.

\section{Approach}

\subsection{Search Strategy for scout robot}

\begin{figure}
\begin{center}
\includegraphics[width=4.2cm]{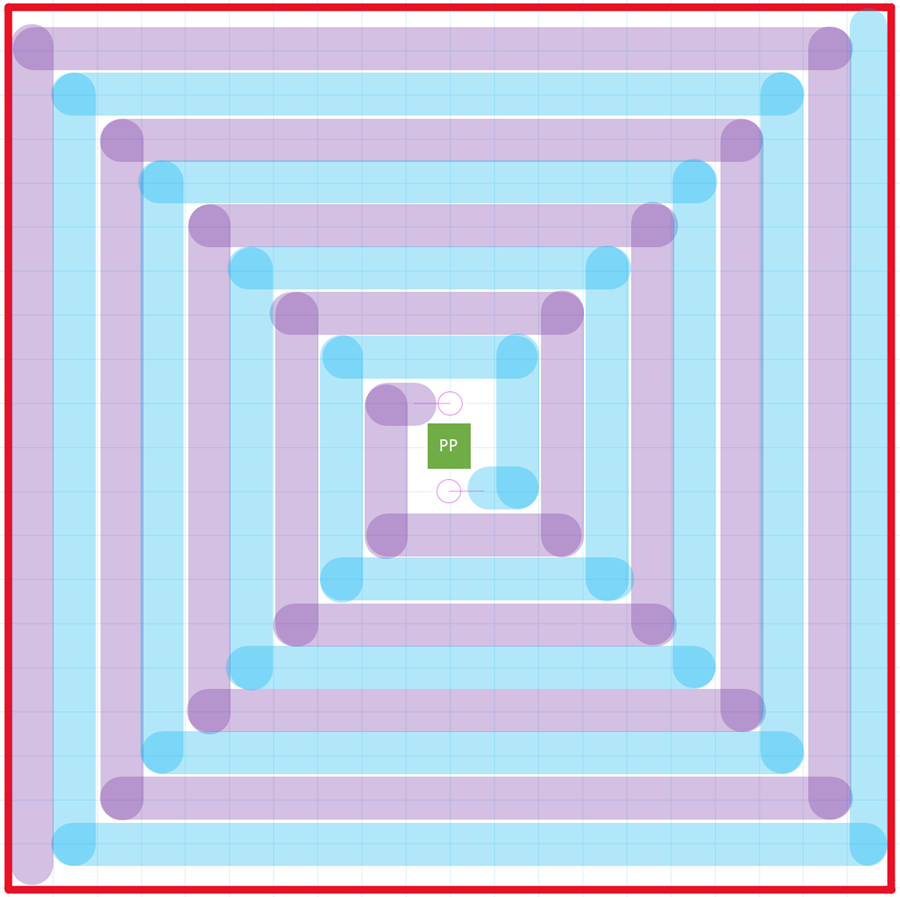}    
\caption{Spiralling outward search pattern} 
\label{fig:spiral_outward}
\end{center}
\end{figure}

 The scout robot can only spot a resource site if it is within its scanning radius. Therefore, the scout robots have to move all over the mining area to find the locations of these resource sites since their locations are unknown. A systematic search strategy is implemented in this research. the mining area is divided into grids and the scout robots move spirally outward. as shown in Fig. \ref{fig:spiral_outward}. This strategy ensures that the resource sites near the processing plant are found first. While spiraling outward, the entire mining area is scanned by the robots efficiently and thoroughly by scanning each grid cell only once with no overlap and there are no missed grid cells to be scanned.

\subsection{Coordination Architecture}

In any lunar mission, the conditions are harsh, and the missions are extremely costly. Co-ordination architecture that provides optimal solutions is always generally preferred, but for lunar missions, better reliability and robustness are preferred so that the work towards the mission keeps going irrespective of many uncertainties. Therefore, a decentralized coordination architecture is chosen for the task allocation system proposed in the thesis. A decentralized coordination architecture will provide a more optimal solution than a distributed coordination architecture and will be more reliable and robust than a centralized coordination where the malfunctioning of the central unit might cause the whole mission to fail. Moreover, a decentralized structure is more flexible and adaptable, thus performing better in unknown and changing environments.

\begin{figure}
\begin{center}
\includegraphics[width=7cm]{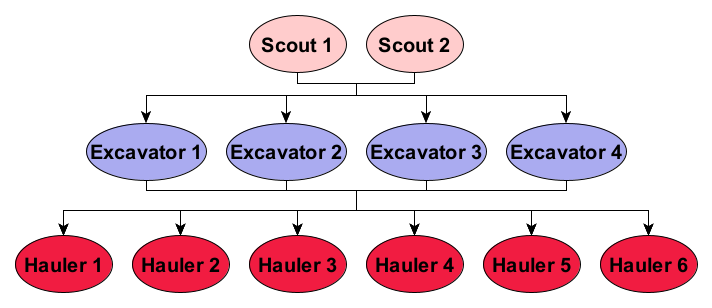}    
\caption{Hierarchically decentralized MRS} 
\label{fig:hierarchical distribution_ifac}
\end{center}
\end{figure}

The tasks to be allocated amongst the heterogeneous MRS are resource localization, excavation, and transport. These tasks are interdependent on each other. A hauler robot cannot transport a resource that has not been excavated by the excavator robot, and the excavator robot cannot excavate resource volatiles from a resource location that has volatiles that have not been found by the Scout robot. Therefore, a resource site needs to be localized first by the scout robot, then the excavator robot needs to excavate resources from that site and then the hauler robot can transport the excavated resources from the site to the processing plant. Due to this kind of interdependency of the tasks and hence the interdependency of the robots, a hierarchically decentralized robot coordination architecture is proposed as shown in Fig. \ref{fig:hierarchical distribution_ifac}.

In this hierarchically decentralized robot coordination architecture, the scout robots allocate tasks to excavator robots, and excavator robots allocate tasks to hauler robots. The scout robot starts performing resource site localization task by moving around in the mining area. Once a resource site is found by the scout robot, it allocates the task of resource excavation to an excavator robot. The excavator robot then starts with the resource excavation task by moving to the resource site and starts digging. The excavator robot needs a hauler robot to transport the excavated resource from the resource site to the processing plant. Therefore, the excavator robot allocates a hauler robot the task of transporting the resource. The hauler robot when it is allocated the task of transporting the resource, navigates to the resource site where the excavator robot that allocated the task is excavating, gets the resources in its bin, and navigates to the processing plant where it empties the bin.

This proposed hierarchically decentralized robot coordination architecture is robust against single-point failures since if a robot fails, another robot of the same type can carry on with the mission. This architecture is also scalable, as there is no strain on one single entity to perform all the processing. The scalability of this architecture is aided by the task allocation system proposed.

\subsection{Multi-Robot Task Allocation}
In this thesis, a variation of MURDOCH by \cite{[7]} for solving the MRTA problem. MURDOCH implements an auctions-based approach for task allocation amongst a group of robots and drives intentional cooperation through these task allocations. MURDOCH, which uses explicit negotiation for its fitness-based auctions for task allocation, has been demonstrated to be an effective and fault-tolerant method for controlling multi-robot systems. The way MURDOCH is formulated makes the muti-robot system scalable, flexible, reliable, and robust. In the following sections, the details of how a variation of MURDOCH is implemented in the group of robots to perform autonomous lunar mining are described.

\subsubsection{Messaging Protocol}
In this method, a broadcast messaging service is used to communicate among robots. As stated by \cite{[7]}, this reduces the communication overhead of the unicast messaging service. A publish/subscribe system of messaging is used for this broadcast messaging service, which is facilitated by ROS$2$.

\subsubsection{Auction Protocol}
Auctions have been proven to be a powerful tool for achieving efficient allocations, even in large-scale environments and a huge group of robot systems, where the acquisition of consistent global state information is difficult or impossible \cite{[7]}. Auctions have been largely studied for their ability to deal with uncertainty. The different types of auctions are discussed in \cite{[8]}. Auctions are well suited for this hierarchically decentralized coordination architecture where the tasks are chronologically dependent. In an auction, the auctioneer does in one way centralize the system, but since anyone in the hierarchy can act as the auctioneer for allocating a task, they still keep the overall coordination hierarchically decentralized. Due to this reason, auctions are highly scalable and are efficient in computation and communication. MURDOCH uses first-price one-round auctions for allocating tasks. The beauty of this type of task allocation system is that a task can be introduced by a human user too by initiating an auction. New robots can be added to the group of robots, and they can take part in auctions and initiate auctions. If some robots fail over the course of time, the other robots of the same type can continue with the task and mission of the system. Thus, the task allocation system used in this research is highly adaptable and flexible.

\subsubsection{Auction Announcement}
A robot that wants to allocate a task to some other robot acts as an auctioneer. The auctioneer robot broadcasts an auction announcement message for the task to be allocated. The broadcast message for an auction announcement by an auctioneer robot has the following format.
    \begin{multline}
        [auctioneer\_robot\_name, task\_type, \\
        task\_location, status = open]
    \end{multline}
A robot can hold multiple auctions at a time for different resource sites it found. To avoid excavator-to-excavator interference, only one excavator robot is allowed to work at a resource site. Therefore, it is less likely that an excavator will hold multiple auctions at a time since it can only start digging once the resource volatile in its bucket is emptied into a hauler robot’s bin.

\subsubsection{Self Utility Evaluation}
In a multi-robot system, there might be more than one robot capable of doing a task that has been announced by an auctioneer robot, and some robots that cannot do the task. Since the auction announcement message is subject-based, and not robot-directed, the message is received by all the robots. The robots that are not capable of doing the task mentioned in the auction announcement message ignore the message. The robots that are capable of doing the task will evaluate their utility to perform the task. Utility is the scalar metric to measure the fitness score of a robot to do the task. Cost is the scalar metric that shows the difficulty of a robot to do the task. In this research, distance from the current location of the robot to the task location is used to determine the cost of that particular robot to perform the task. Distance as a metric for cost is chosen since amongst the capable robots, the difficulty to perform all other functions involved in the task is the same, only the distance from the current location of the robot to the task location will be different. The individual robot’s path planner provides an estimate of the length of the path from the robot’s current location to the task location. All robots have the same map of the mining area, and all robots use RRT-connect as their path planner.
    \begin{equation}
        Cost = Distance
    \end{equation}
    \begin{equation}
        Utility = - Cost
    \end{equation}
Therefore, the longer the distance, the higher the cost, and the lower the utility. A capable robot that is already performing a task tells that the self-utility evaluated is negative infinity, thus claiming that it is currently busy and not available to perform the announced task. The robot’s sensors are noisy, but the robots are honest. Therefore, the self-utility evaluated by the robot is the best estimate, and in action, it can differ based on the obstacles that come in the path of the robot when executing the task.

\subsubsection{Bid submission}
After evaluating their respective utilities, the robots that are capable of performing the announced task submit their respective bids using a broadcast message with the following format.
    \begin{multline}
        [auctioneer\_robot\_name, bidder\_robot\_name, \\
        task\_location, utility]
    \end{multline}   
This broadcast message will be received by all robots, but due to the specific format, it will be recognized as a bid submission message. Thus, the robots that did not announce an auction will ignore the message. The auctioneer that finds its name in these bid submission messages remembers the name of the bidder robot for the specific task location along with its utility. The auctioneer who was holding multiple bids can map the bids based on the specific location mentioned in the message. Moreover, a robot can submit bids for multiple different auctions at a time, and they are differentiated based on the auctioneer robot's name and the task location.

\subsubsection{Winner Selection}
After sufficient time has passed, the auctioneer processes the bids it received from the bid submission messages. The bidder with the highest utility is determined as the winner of the auction by the auctioneer. The highest utility means the least cost path, thus the auctioneer selects the robot as a winner which is closest to the task location. An auctioneer never selects a robot that bids a negative infinity utility since it understands that that particular robot is busy performing another task. Once the winner is selected, the auctioneer broadcasts the winner with an auction winner message format as below.
    \begin{multline}
        [auctioneer\_robot\_name, task\_type, task\_location, \\
        status = open, winner = robot\_name]
    \end{multline}
A robot may win multiple auctions at a time, and the decision to undertake a task is based on the policy that is being implemented. The policies are discussed in the section.

\subsubsection{Closing Auction}
After the winner selection message is broadcast, the auctioneer waits for an acknowledgment message from the winner robot. The acknowledgment message published by the winner robot is formatted as follows.
    \begin{multline}
        [auctioneer\_robot\_name, auction\_winner\_robot\_name, \\
        task\_location, task = accepted/declined]
    \end{multline}
This acknowledgment message notifies the auctioneer if a task is accepted, so that it can close the auction or if a task is declined, to keep the auction open. If a declined acknowledgment message is received by the auctioneer, then the auctioneer chooses the next highest bidder as the winner of the auction and publishes a corresponding message. The auctioneer keeps the auction open until it receives an accepted acknowledgment message from an auction winner robot.

Once the accepted acknowledgment message is received by the auctioneer robot, it sends out an auction closing message in a format as shown below, denoting what task has been allocated to which robot. This auction closing message marks the end of the auction for the task to be allocated, and the robots to whom the tasks have been allocated carry on performing that particular task. The message formats mentioned in these 5 steps can be formulated in a different way too, as long as they can be differentiated from each other.
    \begin{multline}
        [auctioneer\_robot\_name, task\_type, task\_location, \\
        status = closed, task\_allocated\_to\_robot\_name]
    \end{multline}

Three different policies are formulated to work with the previously described auction-based task allocation., which determines how the robots address the auctions.

\subsubsection{Policy 1: First come first serve}
This is the simplest of policies that could be implemented for MRS developed in this thesis. With multiple auctions being held at a given time, this policy dictates that the capable robots only bid in the auction that was initialized first. Unless and until the preceding auction closes, there will be no bids placed for any other auctions.

\subsubsection{Policy 2: Anticipatory coalition formation of excavator and hauler robots}

An excavator robot is not capable of transporting the excavated resource to the processing plant. It requires a hauler robot to do the task of transporting the resource. Therefore, it is safe to say that an excavator robot will always require at least a hauler robot to transport the excavated resource from the resource site to the processing plant. This observation leads to policy $2$ formulation for the MRS of this thesis. Policy 2 dictates that an excavator robot and a hauler robot be paired together. Thus, an excavator will always have a hauler robot to transport its excavated resource. There is no need for an excavator to initialize an auction for its paired hauler to allocate a task. The paired hauler will simply follow its parent excavator from resource site to resource site and transport any resource that is excavated by its parent excavator only. Since there are 4 excavators in the MRS of this thesis, there will be 4 excavator-hauler pairs, and it is called coalition formation of the excavator and hauler robots. There are still $2$ unpaired hauler robots in the system. When an excavator finds its paired hauler busy transporting, it may hold an auction for the excavated resource in which the unpaired haulers can bid. Thus, there are 2 haulers outside of the paired haulers that the excavators can allocate tasks to.

\subsubsection{Policy 3: Serve the nearest one first}

This policy makes use of the information that is available in the MRS. In this policy, when there is more than one auction open at a given time, the capable robots will bid in all the open auctions. If a robot wins in multiple auctions, then the winning robot chooses to do the task which is nearest, since it has the highest utility, thus serving the nearest one first.

\section{Results}

\begin{figure}
\begin{center}
\includegraphics[width=8.4cm]{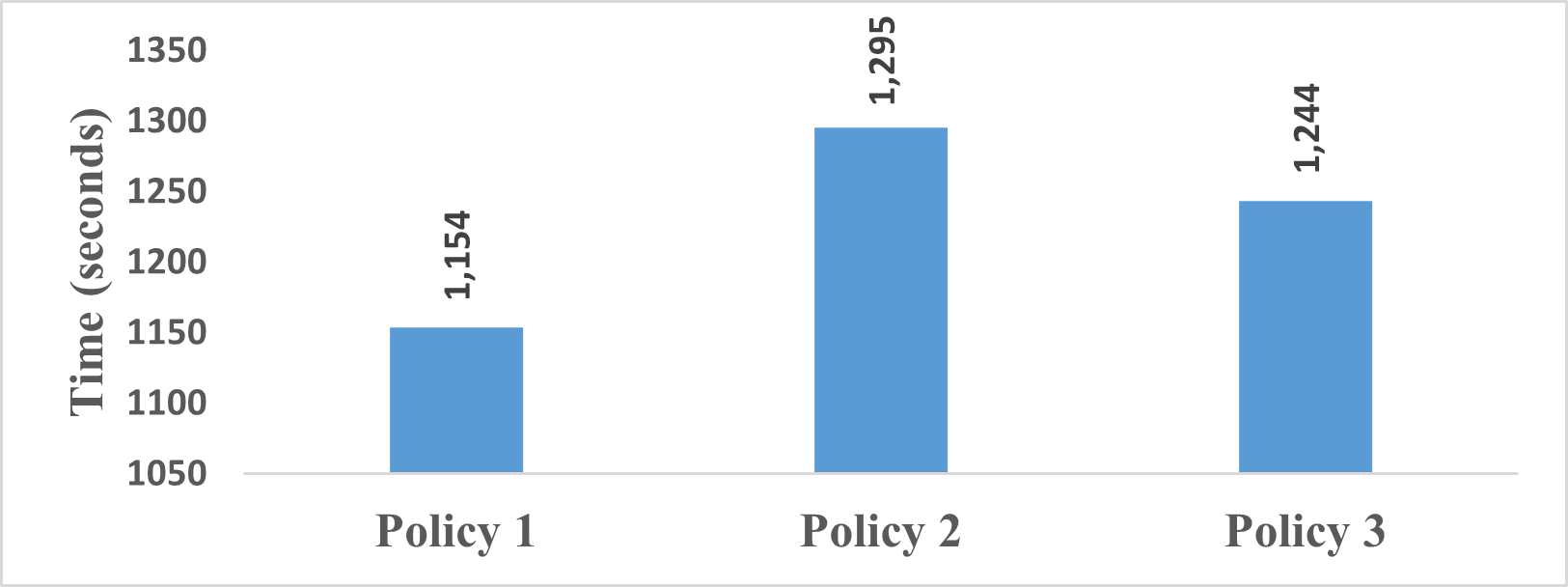}    
\caption{Time taken by each policy for goal completion} 
\label{fig:results1}
\end{center}
\end{figure}

\begin{figure}
\begin{center}
\includegraphics[width=8.4cm]{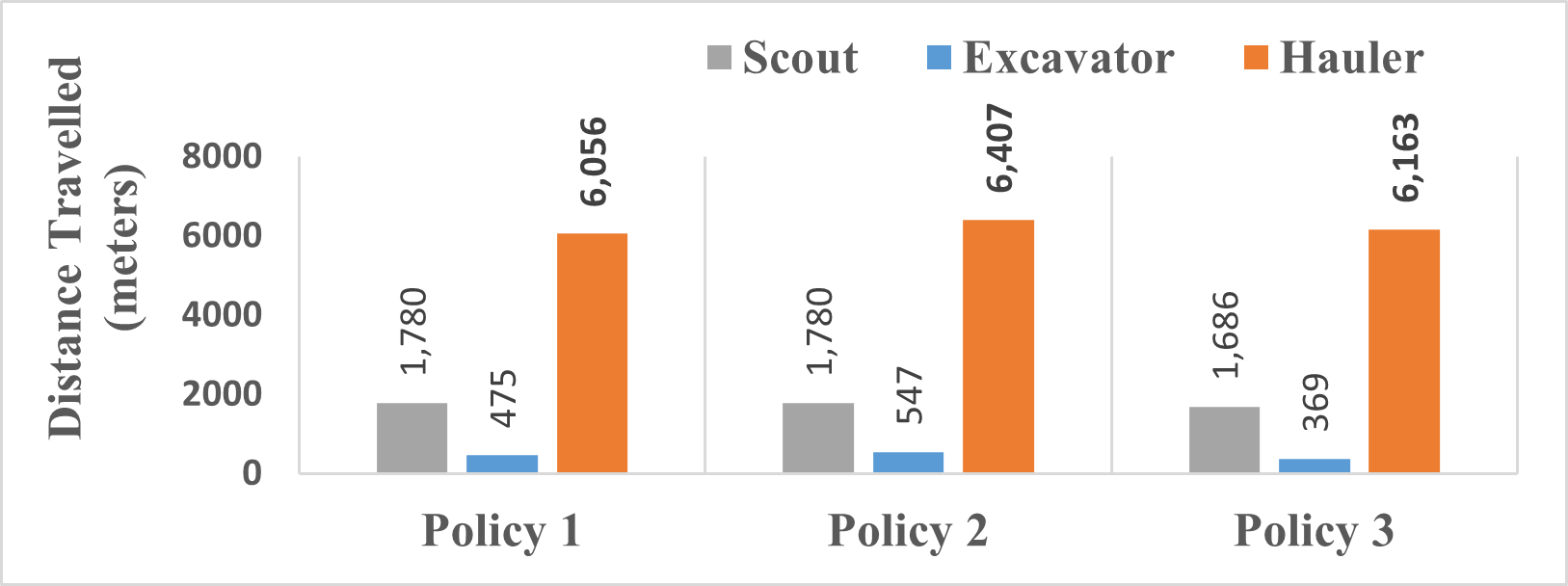}    
\caption{Distance traveled by each kind of robot} 
\label{fig:results2}
\end{center}
\end{figure}

\begin{figure}
\begin{center}
\includegraphics[width=8.4cm]{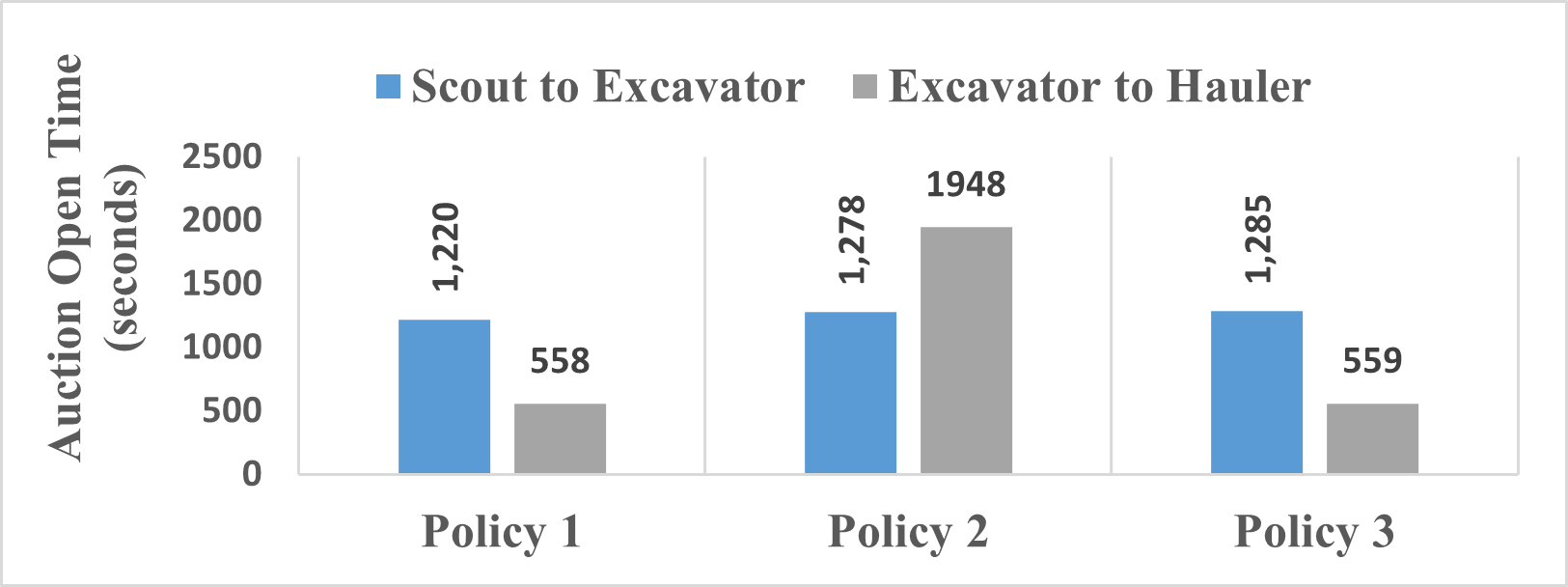}    
\caption{Auction open times for different policies} 
\label{fig:results3}
\end{center}
\end{figure}

Fig. \ref{fig:results1} shows the time taken to find all $10$ resource sites and collect all $64$ mineral objects at the processing plant by the MRS, Fig. \ref{fig:results2} shows the distance traveled by each type of robot and Fig. \ref{fig:results3} shows the time that the auctions were open between scout to excavator and excavator to hauler for the $3$ different policies. It can be seen that policy $1$ took the least amount of time to complete the goal, the distances traveled by excavators were the least in policy $3$, and policy $2$ suffered from the longest auction open times for excavator to hauler task allocations. 

\bibliography{ifacconf}             
                                                   







\end{document}